\pdfoutput=1
\documentclass[11pt]{article}

\usepackage[final]{acl}

\usepackage{times}
\usepackage{latexsym}

\usepackage[T1]{fontenc}
\usepackage[utf8]{inputenc}

\usepackage{microtype}

\usepackage{inconsolata}

\usepackage{graphicx}
\usepackage{multirow}
\usepackage{booktabs}
\usepackage{amsmath}
\usepackage{amssymb}
\usepackage{adjustbox}
\usepackage{tabularx}
\usepackage{hyperref}
\usepackage{makecell}
\usepackage{subcaption}
\usepackage{caption}
\DeclareMathOperator*{\argmax}{arg\,max}

\usepackage{changepage}

\title{Optimising Factual Consistency in Summarisation via Preference Learning from Multiple Imperfect Metrics}

\author{Yuxuan Ye \and Raul Santos-Rodriguez \and Edwin Simpson \\
  Intelligent System Laboratory \\
  University of Bristol \\
  United Kingdom \\
  \texttt{\{yuxuan.ye, enrsr, edwin.simpson\}@bristol.ac.uk} \\}

\begin{document}
\maketitle
\begin{abstract}
%Recent work on language models often applies reinforcement learning with human-annotated preference data to enhance specific capabilities, such as generating informative summaries. 
%However, such data often focuses on overall preferences and overlooks factuality. 
%Since collecting new annotations is costly, we propose to use automatic factuality metrics to obtain factuality preference labels.
%While individual factuality metrics are limited, their combination can effectively capture diverse factual errors. 
%We introduce an automated training pipeline that improves summarisation factuality via preference optimisation. 
%For each source document, we generate lexically similar summary pairs by varying decoding strategies, ensuring the model learns from minor factual errors. 
%To avoid human annotation, we derive preference labels from weak factuality metrics filtering out conflicting cases to improve reliability.
%This results in a high-quality preference dataset constructed with only source documents. 
%Experiments show consistent factuality gains across models, ranging from early encoder-decoder architectures to modern large language models, with smaller models reaching comparable factuality to larger ones.
%Code and data will be released upon acceptance.
%%%%%%%%%%%%%%%%%%%%%%%%%%%%%%%%%%%%%%%%%%%%%%%%%%%%%%%%%%%%%%%%%%

Reinforcement learning with evaluation metrics as rewards is widely used to enhance specific capabilities of language models.
However, for tasks such as factually consistent summarisation, existing metrics remain underdeveloped, limiting their effectiveness as signals for shaping model behaviour.
While individual factuality metrics are unreliable, their combination can more effectively capture diverse factual errors. 
We leverage this insight to introduce an automated training pipeline that improves factual consistency in summaries by aggregating scores from different weak metrics. 
Our approach avoids the need for complex reward shaping by mapping scores to preferences and filtering out cases with high disagreement between metrics. 
For each source document, we generate lexically similar summary pairs by varying decoding strategies, enabling the model to learn from factual differences caused by subtle lexical differences. 
This approach constructs a high-quality preference dataset using only source documents.
Experiments demonstrate consistent factuality gains across models, ranging from early encoder-decoder architectures to modern large language models, with smaller models reaching comparable factuality to larger ones\footnote{Code is available at \url{https://github.com/Haruhi07/MultiMetric}}.

\end{abstract}

\section{Introduction}

\begin{figure*}[ht]
    \centering
    \includegraphics[width=1\linewidth]{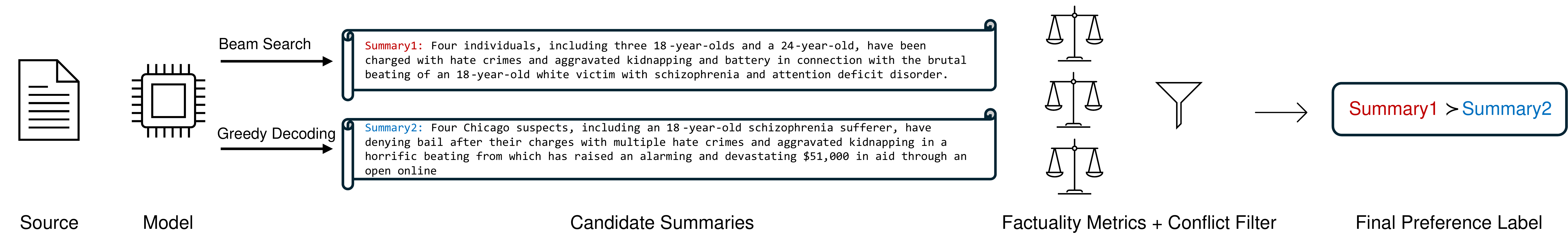}
    \caption{Our method only requires source documents to build a preference dataset.}
    \label{fig:method}
\end{figure*}

Cutting-edge language models have demonstrated impressive capabilities in generating fluent and coherent responses to a wide range of prompts. 
However, maintaining faithfulness and factual consistency remains a persistent challenge, particularly in tasks like summarisation. 
Despite their surface plausibility, model-generated summaries often contain factual inconsistencies or hallucinated details \citep{Huang_2025}.

Recent research has tried to mitigate this issue by incorporating reinforcement learning (RL) to guide models towards more factually consistent outputs.
A critical obstacle lies in designing effective reward signals that can reliably capture and quantify factuality.
Many existing automatic evaluation metrics \citep{lin-2004-rouge,kryscinski-etal-2019-evaluating,zhang-2020-bertscore,laban-2022-summac} have been adopted as reward signals for RL. 
However, even state-of-the-art metrics struggle with subtle inconsistencies and may penalise factually accurate outputs \citep{tang-2023-aggrefact}. 
Furthermore, using a single metric directly as a scalar reward for RL can lead to unstable training, as explored by \citet{roit-2023-factually}.
The training process is influenced by the metric’s reliability and the distribution of reward scores, but the distributions of existing evaluation metrics are not well-studied.
Although combining metrics can broaden error detection coverage \citep{ye-2024-sbertscore}, applying metric combination in RL often requires manual weighting of sub-rewards \citep{gao-2018-april,pasunuru-bansal-2018-multi, wan-bansal-2022-factpegasus, ye-2023-towards}, thus the use of RL is impeded by the complexity of reward shaping.

An alternative is Reinforcement Learning from Human Feedback (\citealp[RLHF]{ouyang-2022-rlhf}), which uses human‑annotated preferences and has proven effective at aligning large language models (LLMs) with broad human values. 
However, recent work \citep{hosking-2024-human, xue-etal-2024-rl} shows that existing RLHF datasets often overlook factuality, despite instructions to annotators to account for it. 
Annotator judgments integrate various considerations, such as trade‑offs among properties, individual biases, and occasional misunderstandings, so the resulting overall preferences can fail to reliably capture factuality. 
Therefore, while constructing a factuality-focused preference dataset is crucial, doing so requires substantial resources and expertise, making scalability a major concern.

To overcome these barriers, this paper proposes a fully automated training pipeline that improves factual consistency in summarisation without relying on human annotations or reference summaries.
We adopt the sampling method from \citet{choi-etal-2024-model}, using the language model itself to generate two summaries by either selecting alternative candidate outputs from the same decoding strategy or using different decoding strategies, as illustrated in Figure \ref{fig:method}. 
In contrast to their work, which paired diverse samples together, our approach ensures that summaries in a pair are lexically similar.
This lexical similarity minimises confounding stylistic or structural differences, allowing the model to focus specifically on factual distinctions, which facilitates the factuality improvement during training. 

With the generated summary pairs, we use an ensemble of factuality metrics to score them and derive preference labels from the scores. 
To mitigate the unreliability of any single metric, we include only those summary pairs for which all selected metrics agree along with preference learning. 
This agreement-based filter removes noisy and contradictory annotations, enhancing the robustness of the preference signal and making the training process more reliable and scalable.

By leveraging lexically similar summary pairs and agreement-based preference labels derived from multiple factuality metrics, our method enables more targeted factuality training than previous RLHF or model-based approaches \citep{stiennon-2020-learning,choi-etal-2024-model}. 
Importantly, in contrast with previous work \citep{roit-2023-factually} that requires an LLM-sized reward model to stabilise the training, we never encounter catastrophic forgetting in our experiments, demonstrating the stability and effectiveness of our proposed method on various language models.
Our method consistently demonstrates factuality improvements on different model architectures, including BART \citep{lewis-2020-bart}, GPT-J \citep{gpt-j}, LLaMA-3 \citep{llama}, and DeepSeek-R1 \citep{deepseek}, showing strong generalisation beyond a single model family or scale. 
Remarkably, our method empowers older and smaller models, such as BART, to achieve factuality performance comparable to that of significantly larger and more recent models, effectively revitalising their potential to produce accurate summaries at lower computational cost.

Our contributions are threefold:
\begin{itemize}
    \item We present a novel, fully automated preference learning pipeline for optimising summarisation factuality, which not only improves LLMs' factuality scores but also elevates the factuality performance of smaller models to the same level.
    \item We show a promising way to adapt multiple existing factuality metrics into training targets. 
    By leveraging preference learning and filtering cases with high disagreement, we improve the reliability of the training data, leading to more robust training in practice. 
    \item We analyse the contribution of lexical similarity between summary pairs and conclude that, 
    with sufficiently accurate preference annotations, 
    similar pairs are more effective for enhancing factuality for summarisers. 
\end{itemize}

\section{Related Work}
\subsection{Factuality Evaluation in Summarisation}
\label{subsec:metrics}
Factuality has become one of the most critical properties to evaluate in recent language models. Depending on the methodologies applied, existing factuality evaluation metrics can be broadly categorised into 3 types.

\paragraph{Similarity-based metrics} Classical similarity-based metrics such as ROUGE \citep{lin-2004-rouge} and BLEU \citep{papineni-2002-bleu} reported the n-gram overlapping ratio between the system and the reference summaries. 
Subsequently, BERTScore 
\citep{zhang-2020-bertscore} replaced the exact word matching with 
embedding-based cosine similarity to enhance robustness to lexical and syntactic variation. 
This idea was then extended to sentence embedding similarity by \citet{ye-2024-sbertscore}, who found that using the source document as reference could shrink the performance gap between similarity-based metrics and other methodologies. 

\paragraph{Question Answering-based metrics} This line of work frames factuality evaluation as a reading comprehension task. 
Key phrases are extracted from the summary, and questions are generated based on their context. 
A question-answering model answers these questions using the source document, then checks whether the answers are consistent with the summary \citep{durmus-2020-feqa, scialom-2021-questeval, fabbri-2022-qafacteval}.
While this approach has shown empirical effectiveness, it usually involves multiple processing stages and models, making it computationally expensive.

\paragraph{Entailment-based metrics} These methods assess whether the source document entails the summary using natural language inference (NLI) models.
Early approaches that simply concatenated the entire document and the summary as input to an NLI model often underperformed.
Recent methods improved performance by segmenting the source document \citep{laban-2022-summac, zha-2023-alignscore} or extracting relational structures for inference \citep{goyal-2020-evaluating, qiu-etal-2024-amrfact}. 
The final factuality score is computed by aggregating the inference results across text segments or extracted relation pairs. 

\subsection{RL for Fine-tuning Language Models}
Reinforcement learning is often applied to fine-tune pre-trained language models, especially for capabilities that are difficult to formalise mathematically.  
Early research introduced interactive or preference learning to define reward functions in RL \citep{gao-2018-april, shapira-2022-interactive}. 
Other previous studies used evaluation metrics as direct reward signals for training \citep{pasunuru-bansal-2018-multi, ye-2023-towards}, but these approaches often suffered from distribution shift and required careful reward design to prevent catastrophic forgetting and to combine multiple, sometimes contradictory, reward components.

With the advent of LLMs, RL has been widely used with human feedback to enforce desirable properties such as safety, which are difficult to guarantee through supervised fine-tuning alone \citep{llama}. 
More recently, DeepSeek-R1 have demonstrated that RL can also facilitate emergent capabilities, such as reasoning \citep{deepseek}. 
However, this depends on sparse rule-based rewards that may be difficult to learn from. 
While RLHF \citep{ouyang-2022-rlhf} can tune the model for properties that are hard to define, the annotators make an overall judgment that might ignore factual errors \citep{hosking-2024-human}, leading to underperformance on factuality \citep{wang-2024-factual,augenstein-2023-factuality}.

To avoid the limitations and costs of human annotation, \citet{choi-etal-2024-model} proposed to label summary pairs using simple rules. 
It over-simplifies the problem and can introduce noise into the labels. 
Therefore, we propose instead to use a combination of existing evaluation metrics that directly target factual consistency. Our experiments suggested solid gains in factuality compared to their approach. 

\section{Methods}
\subsection{Summary Generation}
\label{subsec:generation}
Given a source document $\mathbf{x}$, different decoding strategies can lead to various outputs $\mathbf{y}$.

\paragraph{Beam Search} selects the top-$k$ most likely partial sequences at each timestep $t$, by extending each of the $k$ token sequences from the previous timestep, $\mathbf{y}_{<t}$, with all possible tokens. Each sequence is scored by its log probability conditioned on  the source document $\mathbf{x}$.  The hyperparameter $k$ is known as the beam size. The output $\mathbf{y}_{beam}$ with length $L$ can be expressed as:
\begin{equation}
    \mathbf{y}_{beam} = \argmax_{\mathbf{y} \in B}\sum_{t=1}^{L}{\log{P(y_t|\mathbf{y}_{<t},\mathbf{x})}}
    \label{eq:beam}
\end{equation}
where  $B$ is the set of top-$k$ candidate sequences identified during decoding.

\paragraph{Greedy Decoding} chooses the most likely token at each timestep:
\begin{equation}
    y_{t}=\argmax_{y_t}\log{P(y_t|\mathbf{y}_{<t},\mathbf{x})}
\end{equation}

\paragraph{Random Sampling} samples each token from the vocabulary's probability distribution at each timestep. 
The distributions are derived from logits using the softmax function:
\begin{equation}
    y_{t}\sim \mathrm{softmax}\left(\frac{LM(y_t|\mathbf{y}_{<t},x)}{\tau}\right)
\end{equation}
where $LM(\cdot)$ denotes the logit output of each timestep, and temperature $\tau$ controls the sampling distribution. 
A higher $\tau$ increases diversity by adding more variance to the outputs.

Recent LLMs often employ the sampling-based decoding strategies to enhance output diversity \citep{llama, deepseek}. 
Prior research has shown that beam search tends to yield higher factuality scores compared to other decoding strategies, especially random sampling \citep{wan-2023-faithfulness,choi-etal-2024-model}. 
In contrast, greedy decoding generally produces outputs that are lexically similar but less factually consistent than beam search outputs, as it is biased towards locally optimal token choices.

In this paper, we aim to train a model to avoid generating highly probable but factually inconsistent summaries. 
To do this, we can generate pairs of summaries with minimal differences from the same decoding strategy. 
For example, we can take the second most probable sequence produced by beam search as follows, where $\mathbf{y}_{beam}$ is the standard beam search output from Equation \ref{eq:beam}.

\begin{equation}
    \mathbf{y}_{beam'} = \argmax_{\mathbf{y}\neq \mathbf{y}_{beam},\mathbf{y}\in B}\sum_{t=1}^{L}{\log{P(y_t|\mathbf{y}_{<t},x)}}
    \label{eq:alternative}
\end{equation}

This ensures that $\mathbf{y}_{beam}$ and $\mathbf{y}_{beam'}$ differ only slightly, enabling the evaluation metrics to focus on factuality differences, rather than stylistic or structural variations that could bias the evaluation.

\subsection{Data Annotation}
In this subsection, we leverage multiple factuality metrics to score summaries generated in the previous step. 
Prior research \citep{choi-etal-2024-model} used a heuristic to identify target summaries, rather than scoring each one, where beam search-generated summaries were always selected as the winning completions in preference learning.
This introduces noise into the training data: 
it assumes that the higher average factuality score of beam search necessarily corresponds to more factual summaries individually, 
but it struggles when beam search and greedy decoding produce similar outputs, in which cases the greedy decoding could be more accurate. 

To address this issue, instead of over-trusting beam search-generated summaries, we use multiple weak factuality metrics to score the summaries and derive preference labels from them. 
Since scores from different metrics are not directly comparable, we convert these heterogeneous scores to binary preference labels so that they can be aggregated. 
Then we employ a conflict resolution strategy to filter out inconsistent preference labels. 
The annotation process works as follows:
\begin{enumerate}
    \item For each metric $m_i$, we obtain score $S_{m_i}(\mathbf{y,x})$ for summary $\mathbf{y}$ given source $\mathbf{x}$.
    \item For each pair of summaries $(\mathbf{y}_1,\mathbf{y}_2)$ related to the same source document $\mathbf{x}$, we obtain its binary preference label under the metric $m$, which can be written as $P_{m_i}(\mathbf{y}_1,\mathbf{y}_2,\mathbf{x}) = \operatorname{sign}(S_{m_i}(\mathbf{y}_1,\mathbf{x}) - S_{m_i}(\mathbf{y}_2,\mathbf{x}))$
    \item The conflict filter checks $\{P_{m_i}(\mathbf{y}_1,\mathbf{y}_2,\mathbf{x})\}$ and only keeps the data with consistent preference labels under all metrics $m_i$.
\end{enumerate}

\subsection{Training with DPO}
Using the preference data obtained from the previous step, we apply Direct Preference Optimization (\citealp[DPO]{rafailov-2023-direct}) to train the language models towards improved factuality.
Compared to RL, DPO directly optimises models without requiring a separate reward model, reducing complexity and improving training efficiency.
Given summary pairs with corresponding preference labels, DPO adjusts the model parameters to increase the likelihood of generating the preferred summary. 
The loss function of DPO can be written as:
\begin{equation*}
\begin{aligned}
    &L(\theta)=&\\
    &\mathbb{E}_{(\mathbf{x},\mathbf{y}_{\{w,l\}})}[\log{\sigma(\beta(f_{\theta}(\mathbf{x},\mathbf{y}_w)-f_{\theta}(\mathbf{x},\mathbf{y}_l)))}]&
\end{aligned}
\end{equation*}
where $\sigma$ is the sigmoid function, 
$f$ is the log probability that the model assigns to a summary,
$\theta$ represents the model parameters to optimise, 
$\beta$ is a temperature parameter,
and $\mathbf{y}_{\{w,l\}}$ denote the winning and losing summaries in the pair, respectively.

\section{Experiments}
\subsection{Experimental Setup and Implementations}

\subsubsection{Dataset and Evaluation Metrics}
To ensure consistency with prior work \citep{choi-etal-2024-model}, we evaluate our approach on XSUM \citep{narayan-2018-dont} and TL;DR \citep{volsketal-2017-tldr}. 
Both datasets require the summarisation of long news articles or Reddit posts into single-sentence summaries, posing challenges for the summarisers to identify key information and assemble it correctly.
Table \ref{tab:dataset} presents the characteristics of the two datasets. Numbers in parentheses refer to the test split while other numbers are for the training split. Length refers to the total number of words in the text. Compression Ratio is computed between source length and summary length.

\begin{table}[h]
    \centering
    \scriptsize
    \begin{tabular*}{\columnwidth}{@{}@{\extracolsep{\fill}}ccccc@{}}
    \toprule
     \textbf{Dataset} & \textbf{Size} &\textbf{\makecell[c]{Source\\Length}} & \textbf{\makecell[c]{Summary\\Length}} & \textbf{\makecell[c]{Compression\\Rate}} \\
    \midrule
     XSUM & 204045(11334) & 430(433) & 23(23) & 5.35\%(5.31\%) \\ 
     TL;DR & 116722(6553) & 313(314) & 31(31) & 9.90\%(9.87\%) \\ 
    \bottomrule
    \end{tabular*}
    \caption{Characteristics of XSUM and TL;DR datasets.}
    \label{tab:dataset}
\end{table}

For evaluation, we apply the same automatic metrics as in the previous work \citep{choi-etal-2024-model} to ensure a fair comparison. 
AlignScore \citep{zha-2023-alignscore} and FactCC \citep{kryscinski-etal-2019-evaluating} reflect factuality, while ROUGE-L \citep{lin-2004-rouge} and BARTScore \citep{yuan-etal-2021-bartscore} check the coherence.
The definitions of these metrics are in Appendix \ref{app:metrics}. 
In addition, we employ ChatGPT to compare our approach against the baselines as LLMs have shown promising results in directly evaluating generative tasks \citep{gekhman-2023-trueteacher,luo-2023-evaluator}. 
We further analyse shifts in common types of factual consistency error  to understand the impact of our training pipeline, again using ChatGPT to categorise mistakes.

\subsubsection{Language Model Selection}
To demonstrate the robustness of our method, we select a variety of language models with different scales and capabilities. 
Model specifications are listed in Table \ref{tab:lm_spec}. 
We select BART-large \citep{lewis-2020-bart} to represent encoder-decoder models that were widely employed before the advent of LLMs. 
We select GPT-J-6B \citep{gpt-j}, LLaMA-3.2-3B \citep{llama}, and DeepSeek-R1-Distill-Qwen-7B \citep{deepseek} as they are representative LLMs trained for different purposes. 
Due to their large sizes, we apply LoRA \citep{hu-2021-lora} and only train an adapter during fine-tuning.

\begin{table}[htbp]
    \centering
    \resizebox{\columnwidth}{!}{
    \begin{tabular}{@{}@{\extracolsep{\fill}}cccccc@{}}
    \toprule
    \multirow{2}{*}{\textbf{Model}} & \multirow{2}{*}{\textbf{Size}} & \multirow{2}{*}{\textbf{Architecture}} & \multirow{2}{*}{\textbf{\makecell[c]{Pre-release\\Fine-tuning}}} & \multirow{2}{*}{\textbf{\makecell[c]{Main\\Ability}}} & \multirow{2}{*}{\textbf{\makecell[c]{Fine-tuning\\Scale}}} \\
    &&&&&\\
    \midrule
    \multirow{2}{*}{BART-large} & \multirow{2}{*}{406M} & \multirow{2}{*}{Encoder-Decoder} & \multirow{2}{*}{SFT} & \multirow{2}{*}{Summarisation} & \multirow{2}{*}{Full}\\ 
    &&&&&\\
    \midrule
    \multirow{2}{*}{GPT-J} & \multirow{2}{*}{6B} & \multirow{2}{*}{Decoder} & \multirow{2}{*}{SFT} & \multirow{2}{*}{\makecell[c]{Open-ended\\Generation}} & \multirow{2}{*}{Adapter}\\ 
    &&&&&\\
    \midrule
    \multirow{2}{*}{LLaMA-3.2} & \multirow{2}{*}{3B} & \multirow{2}{*}{Decoder} & \multirow{2}{*}{SFT+RL} & \multirow{2}{*}{Instruction} & \multirow{2}{*}{Adapter} \\ 
    \\
    \midrule
    \multirow{2}{*}{\makecell[c]{DeepSeek-R1\\(Distill-Qwen)}} & \multirow{2}{*}{7B} & \multirow{2}{*}{Decoder} & \multirow{2}{*}{SFT+RL} & \multirow{2}{*}{Reasoning} & \multirow{2}{*}{Adapter} \\ 
    &&&&&\\
    \bottomrule
    \end{tabular}
    }
    \caption{Specifications of the selected language models.}
    \label{tab:lm_spec}
\end{table}

\paragraph{GPT-J} is an alternative for GPT-3 \citep{gpt3} and was only tuned with SFT. 
It can perform specific tasks given a prompt but it is suggested to apply task-oriented SFT beforehand. 

\paragraph{LLaMA-3.2} utilised RL during its training process, specifically through RLHF, to enhance its alignment with human preferences and improve the quality of its responses.

\paragraph{DeepSeek-R1} is a mixture-of-experts model with 671B parameters, providing impressive reasoning ability on a wide range of tasks including maths and coding. In this paper, we use its distilled model based on Qwen2.5 \citep{qwen2.5} to balance the training efficiency and reasoning quality.

For GPT-J, SFT is required before RL, so we only use a simple prompt as it will learn to summarise during SFT.
For LLaMA and DeepSeek, we avoid fine-tuning them on specific tasks before applying RL, simulating real-world conditions where they are provided only with task instructions.
To maintain consistency across experiments, we use the same generic summarisation prompt for all LLMs.
Details of the prompt are available in Appendix \ref{app:prompt}, along with the processing steps for DeepSeek's chain-of-thought output.

\subsubsection{Decoding Strategies}
As highlighted in prior studies \citep{holtzman-2019-curious, choi-etal-2024-model}, decoding strategies can impact factuality. 
In this section, we study how decoding strategies influence factual consistency on our selected datasets and select which to use in the consequent experiments.
\begin{table}[ht]
    \centering
    \tiny
    \resizebox{\columnwidth}{!}{
    \begin{tabular}{ccccccc}
    \toprule
     \multirow{2}{*}{\textbf{Dataset}} & \multirow{2}{*}{\textbf{Model}} & \multicolumn{5}{c}{\textbf{AlignScore($\uparrow$)}} \\
    \cmidrule{3-7}
     &  & BS\#1 & BS\#2 & RS\#1 & RS\#2 & Greedy \\
    \midrule
    \multirow{4}{*}{\rotatebox{90}{XSUM}} & BART & \textbf{61.9} & 61.5 & 19.2 & 18.4 & 58.9 \\ 
     & GPT-J & \textbf{59.7} & 58.3 & 17.4 & 17.3 & 50.5 \\ 
     & LLaMA & \textbf{86.1} & 85.3 & 67.3 & 66.5 & 83.6 \\ 
     & DeepSeek & \textbf{82.5} & 82.4 & 60.2 & 59.6 & 80.5 \\ 
    \midrule
    \multirow{4}{*}{\rotatebox{90}{TL;DR}} & BART & \textbf{84.9} & 84.7 & 42.5 & 41.0 & 80.6 \\ 
     & GPT-J & \textbf{89.6} & 89.0 & 60.3 & 60.2 & 83.6 \\ 
     & LLaMA & \textbf{91.4} & 90.6 & 83.7 & 83.6 & 90.7 \\ 
     & DeepSeek & \textbf{89.1} & 88.9 & 75.6 & 75.8 & 87.9 \\ 
    \bottomrule
    \end{tabular}
    }
    \caption{AlignScore of different decoding strategies.}
    \label{tab:decoding_score}
\end{table}

From Table \ref{tab:decoding_score}, we observe that the first candidate from beam search (BS\#1) consistently outperforms other decoding strategies, including greedy decoding and random sampling (RS\#1 and RS\#2). 
The latter strategies introduce excessive randomness or focus too narrowly on local token probabilities, leading to lower factuality. 
Therefore, in our experiments, we primarily use beam search and greedy decoding, as these strategies provide relatively high factual accuracy while the mix of strategies allows us to generate different summaries for the same source.
For final evaluation, we use the first beam search output to ensure the highest factuality.

\subsection{Factuality Scoring Metrics}
Among the metrics mentioned in \ref{subsec:metrics}, we utilise SBERTScore \citep{ye-2024-sbertscore} and SummaC-Conv \citep{laban-2022-summac}, representing similarity-based and NLI-based metrics respectively.
Their definitions are in Appendix \ref{app:def}.
These metrics, while slightly less powerful than state-of-the-art alternatives, are more computationally efficient. 
We exclude QA-based metrics not only due to their high computational cost, but also because they require a question generation model trained on the same dataset, which is not available for TL;DR.

\subsection{Baselines}

We compare our proposed approach with three baselines: supervised fine-tuning (SFT), reinforcement learning from human feedback (RLHF), and model-based preference optimisation (\citealp[MPO]{choi-etal-2024-model}).
Both SFT and RLHF are common fine-tuning methods that rely on either golden references or human annotations. 
SFT trains on reference summaries, while RLHF is often applied after SFT, using human preference rankings to optimise via RL rather than direct supervision.

We reuse the official SFT checkpoints for BART on XSUM\footnote{\url{https://huggingface.co/facebook/bart-large-xsum}} and RLHF checkpoints for GPT-J\footnote{\url{https://huggingface.co/CarperAI/openai\_summarize\_tldr\_ppo}} on TL;DR.
For RLHF on the other models, we perform training using the TRL\footnote{\url{https://huggingface.co/docs/trl/main/en/ppo\_trainer}} library, with the trl-lib/tldr-preference dataset\footnote{\url{https://huggingface.co/datasets/trl-lib/tldr-preference}}, which includes preference labels based on overall human judgments that are not specifically focused on factuality.

MPO \citep{choi-etal-2024-model} avoids the need to score summaries by assuming that beam search-generated summaries 
are more factually consistent than those generated by other decoding strategies.
However, while beam-search generates more factual summaries on average, individual summaries are not guaranteed to be the most factually consistent, leading to some mislabelled pairs.
This resulted in huge performance degradation for MPO when applied to similar summary pairs in their study.
Our proposed method overcomes this by using multiple computationally efficient metrics to annotate generated summaries, allowing greater resilience to input similarity and better utilisation of summaries from various decoding strategies.

\subsection{Experimental Results}

We built two datasets for each model, using (BS\#1,BS\#2) and (BS\#1,Greedy) respectively.
Table \ref{tab:new_main_tab} presents the better factuality performance between the two settings for MPO and our approach individually. 
We do not report RLHF results for XSUM due to the lack of a human preference dataset, nor do we include DeepSeek RLHF results for TL;DR, as we cannot learn a reward model for it on a preference dataset without chain-of-thought examples.

\begin{table}[htbp]
    \renewcommand{\arraystretch}{1.3}
    \centering
    \resizebox{\columnwidth}{!}{
        \begin{tabular}{ccccccc}
        \toprule
        \textbf{Dataset} & \textbf{Model} & \textbf{Method} & \textbf{AlignScore($\Delta$)} & \textbf{BARTScore} & \textbf{FactCC} & \textbf{ROUGE-L} \\
        \midrule
        \multirow{12}{*}{\rotatebox{90}{XSUM}} & \multirow{3}{*}{BART} & SFT & 61.9(\textbackslash) & -3.69 & 65.7 & \textbf{36.4} \\ 
        & & MPO & 62.0(+0.1) & -3.67 & 76.0 & 33.5 \\ 
        & & Ours & \textbf{86.6}(\textbf{+24.7}) & \textbf{-3.63} & \textbf{83.3} & 33.9 \\ 
        \cline{2-7}
        & \multirow{3}{*}{GPT-J} & SFT & 59.7(\textbackslash) & -3.63 & 62.4 & \textbf{25.0} \\ 
        & & MPO & 53.5(-6.2) & -3.90 & 73.8 & 23.6 \\ 
        & & Ours & \textbf{75.8(+16.1)} & \textbf{-3.57} & \textbf{80.6} & 22.3 \\ 
        \cline{2-7}
        & \multirow{3}{*}{LLaMA} & SFT & 86.1(\textbackslash) & -3.48 & 75.1 & \textbf{19.2} \\ 
        & & MPO & 79.8(-6.3) & -3.49 & 76.0 & 18.8 \\ 
        & & Ours & \textbf{88.7(+2.6)} & \textbf{-3.47} & \textbf{77.4} & 18.3 \\ 
        \cline{2-7}
        & \multirow{3}{*}{DeepSeek} & SFT & 82.5(\textbackslash) & -3.34 & 75.9 & \textbf{14.8} \\ 
        & & MPO & 81.3(-1.2) & -3.39 & 78.8 & 12.5 \\ 
        & & Ours & \textbf{83.2(+0.7)} & \textbf{-3.23} & \textbf{80.5} & 14.0 \\ 
        \midrule
        \multirow{15}{*}{\rotatebox{90}{TL;DR}} & \multirow{4}{*}{BART} & SFT & 84.9(\textbackslash) & -3.48 & 43.2 & \textbf{25.8} \\
        & & RLHF & 73.1(-11.8) & -3.95 & 41.5 & 22.6 \\
        & & MPO & 88.1(+3.2) & -3.40 & 56.9 & 24.2 \\
        & & Ours & \textbf{94.2(+9.3)} & \textbf{-3.33} & \textbf{65.6} & 22.4 \\
        \cline{2-7}
        & \multirow{4}{*}{GPT-J} & SFT & 89.6(\textbackslash) & -3.69 & 30.9 & \textbf{26.8} \\
        & & RLHF & 81.5(-8.1) & -3.59 & 34.1 & 23.4 \\
        & & MPO & 92.3(+2.7) & -3.53 & 37.5 & 23.7 \\
        & & Ours & \textbf{93.8(+4.2)} & \textbf{-3.44} & \textbf{46.0} & 22.3 \\
        \cline{2-7}
        & \multirow{4}{*}{LLaMA} & SFT & 91.4(\textbackslash) & -3.81 & 73.7 & 15.6 \\
        & & RLHF & 90.2(-1.2) & -3.78 & 64.1 & \textbf{18.3} \\
        & & MPO & 86.4(-5.0) & -3.78 & 81.3 & 15.4 \\
        & & Ours & \textbf{93.5(+2.1)} & \textbf{-3.74} & \textbf{84.1} & 15.1 \\
        \cline{2-7}
        & \multirow{3}{*}{DeepSeek} & SFT & 89.1(\textbackslash) & -3.79 & 66.7 & \textbf{15.8} \\
        & & MPO & 89.7(+0.6) & -3.77 & 72.5 & 15.1 \\
        & & Ours & \textbf{90.9(+1.8)} & \textbf{-3.69} & \textbf{75.8} & 15.1 \\
        \bottomrule
        \end{tabular}
    }
    \caption{Evaluation results on the two datasets. $\Delta$ refers to the performance difference over SFT results. The best results are highlighted in \textbf{bold}.}
    \label{tab:new_main_tab}
\end{table}

Our approach consistently outperforms all three baselines on AlignScore, FactCC and BARTScore, bringing positive effects to all models across both datasets, and the largest improvements across all models. 
RLHF and MPO sometimes decreased the factuality, specifically for LLaMA on both datasets.

For ROUGE-L, we found the same trade-off between it and the factuality performance as in \citet{choi-etal-2024-model}. 
ROUGE is computed between the generated summary and the reference summary, which is directly used for SFT.
Note that a previous study \citep{maynez-etal-2020-faithfulness} has indicated that some human written reference summaries contain hallucinations. 
Considering the large factuality improvement obtained from our approach, we think this trade-off is within the acceptable range.

\begin{table}[htbp]
    \centering
    \tiny
    \begin{tabular*}{0.9\columnwidth}{@{}@{\extracolsep{\fill}}ccccc@{}}
    \toprule
     \multirow{2}{*}{\textbf{Dataset}} & \multirow{2}{*}{\textbf{Model}} & \multicolumn{3}{c}{\textbf{Baseline}} \\
    \cmidrule{3-5}
     &  & SFT & RLHF & MPO \\
    \midrule
    \multirow{4}{*}{XSUM} & BART & 51.4 & \textbackslash & 52.0 \\ 
     & GPT-J & 44.2 & \textbackslash & 80.0 \\ 
     & LLaMA & 42.0 & \textbackslash & 54.0 \\ 
     & DeepSeek & 39.0 & \textbackslash & 52.4 \\ 
    \midrule
    \multirow{4}{*}{TL;DR} & BART & 47.2 & 40.4 & 54.8 \\ 
     & GPT-J & 46.8 & 42.8 & 61.6 \\ 
     & LLaMA & 43.4 & 39.2 & 74.6 \\ 
     & DeepSeek & 40.8 & \textbackslash & 58.6 \\ 
    \bottomrule
    \end{tabular*}
    \caption{The win rates against baselines, judged by ChatGPT for overall summary quality.}
    \label{tab:gpt_eval}
\end{table}

The results show that our approach is more effective at improving summary factuality compared to RLHF on human-labelled datasets or MPO's heuristic preference label generation, while maintaining overall quality compared to the reference summaries used by SFT. 
This highlights the benefit of scoring summaries based on factuality metrics rather than relying on heuristic preferences.

Across the four models, BART gained the largest improvement with an AlignScore increase of $24.7$ on XSUM and $9.3$ on TL;DR. 
Although LLMs had less headroom for the factuality improvement, our method still managed to increase their scores marginally.
It is worth noting that our training pipeline sealed the gap between BART and the LLMs and led to better post-training performance, making it possible to 
apply BART where computing resources are limited.
The DeepSeek reasoning model received the least improvement. 
We speculate that this is because our preference labels are only decided by the final summary, so errors made in the thinking process generated before it would be overlooked by the scoring metrics, resulting in a noisy training signal. 

\subsection{Overall Quality Evaluation}

To gain a better understanding on the overall quality of the generated summaries, we use ChatGPT-4o-mini to evaluate them based on not just factuality, but also informativeness, coherence, and legibility. 
We randomly selected 500 source documents from each dataset, applied different models to generate summaries and asked ChatGPT to compare them in pairs.
The full evaluation prompt can be found in Appendix \ref{app:prompt}.
We compared the summaries from our approach against those from the baselines. 
Some win rates against RLHF are not available due to the availability of the human preference dataset.

Table \ref{tab:gpt_eval} shows that our summaries were preferred over MPO but less preferred than SFT summaries. 
This is likely because SFT directly trains on human-written reference summaries, while ours focus on factuality, leading to potentially less fluency or informativeness. 
RLHF summaries are also more preferred because they are originally trained to align with human values, thus being more likely to be selected by ChatGPT, which has also been trained with the same purpose.
However, previous discussion has confirmed the competitive overall quality of our summaries.
Therefore, we asked ChatGPT to output the reasons for its selections, and found out that the preferred summaries contained excessive details, while our summaries are more abstract and discarded some of the unnecessary details to reduce the risk of generating inconsistent content (Appendix \ref{app:reason}).
This suggests a trade-off between factual consistency and summary style, which aligns with previous findings \cite{hosking-2024-human} that overall judgements may neglect factuality. 

\section{Analysis}
\subsection{Ablation Study}

We studied the effectiveness of each component in our approach and present their influence in Table \ref{tab:main_tab}. 
Introducing a single factuality metric to score the summary did not always lead to improvements. 
For example, when only one metric was applied, LLaMA and DeepSeek occasionally showed decreased factuality scores.
However, when multiple factuality metrics were applied, all models showed improvement. 
Additionally, filtering out inconsistent labels further enhanced performance, likely because contradicting labels may appear in different batches, thereby adding noise during training. 

\begin{table}[htbp]
    \renewcommand{\arraystretch}{1.5}
    \centering
    \resizebox{\columnwidth}{!}{
    \begin{tabular}{ccccccccc}
    \toprule
     \multirow{3}{*}{\textbf{Dataset}} & \multirow{3}{*}{\textbf{Model}} & \multirow{3}{*}{\textbf{\makecell[c]{Decoding\\Strategy}}} & \multirow{3}{*}{\textbf{\makecell[c]{Pair\\Similarity}}} & \multicolumn{4}{c}{\textbf{Scoring Metric}} & \multirow{3}{*}{\textbf{\makecell[c]{SFT\\Results}}} \\
    \cline{5-8}
     & & & & SBERT & SummaC & \makecell[l]{\enspace SBERT\\+SummaC} & \makecell[l]{\enspace SBERT\\+SummaC\\+Filter} & \\
    \midrule
    \multirow{8}{*}{\rotatebox{90}{XSUM}} & \multirow{2}{*}{BART} & (BS\#1,BS\#2) & 0.940 & 71.4 & 79.7 & 78.5 & \textbf{86.6} & \multirow{2}{*}{61.9}\\ 
     &  & (BS\#1,Greedy) & 0.826 & 75.0 & 81.7 & 79.9 & \textbf{86.1} & \\ 
    \cline{2-9}
     & \multirow{2}{*}{GPT-J} & (BS\#1,BS\#2) & 0.973 & 60.0 & 54.1 & \textbf{71.7} & 70.9 & \multirow{2}{*}{59.7} \\ 
     &  & (BS\#1,Greedy) & 0.773 & 68.2 & 73.9 & 70.0 & \textbf{75.8} & \\ 
    \cline{2-9}
     & \multirow{2}{*}{LLaMA}& (BS\#1,BS\#2) & 0.938 & 85.0 & 86.5 & 87.5 & \textbf{88.7} & \multirow{2}{*}{86.1} \\ 
     &  & (BS\#1,Greedy) & 0.889 & 85.5 & 84.3 & 86.3 & \textbf{87.1} & \\ 
    \cline{2-9}
     & \multirow{2}{*}{DeepSeek}& (BS\#1,BS\#2) & 0.985 & 81.1 & 82.6 & 82.8 & \textbf{83.0} & \multirow{2}{*}{82.5} \\ 
     &  & (BS\#1,Greedy) & 0.843 & 80.7 & 82.2 & 83.1 & \textbf{83.2} & \\ 
    \midrule
    \multirow{8}{*}{\rotatebox{90}{TL;DR}} & \multirow{2}{*}{BART} & (BS\#1,BS\#2) & 0.954 & 94.0 & 91.3 & \textbf{94.7} & 94.1 & \multirow{2}{*}{84.9} \\ 
     &  & (BS\#1,Greedy)& 0.802 & 93.1 & 91.3 & \textbf{94.4} & 94.2 &  \\ 
    \cline{2-9}
     & \multirow{2}{*}{GPT-J} & (BS\#1,BS\#2) & 0.943 & 92.9 & 95.3 & \textbf{95.6} & 93.7 & \multirow{2}{*}{89.6} \\ 
     &  & (BS\#1,Greedy) & 0.751 & 91.9 & 91.6 & \textbf{94.2} & 93.8 & \\ 
    \cline{2-9}
     & \multirow{2}{*}{LLaMA}& (BS\#1,BS\#2) & 0.909 & 92.1 & 90.8 & 91.8 & \textbf{93.5} & \multirow{2}{*}{91.4} \\ 
     &  & (BS\#1,Greedy) & 0.868 & 89.9 & 91.0 & 91.5 & \textbf{92.9} & \\ 
    \cline{2-9}
     & \multirow{2}{*}{DeepSeek}& (BS\#1,BS\#2) & 0.972 & 88.7 & 85.6 & 89.2 & \textbf{90.9} & \multirow{2}{*}{89.1} \\ 
     &  & (BS\#1,Greedy) & 0.735 & 89.5 & 88.8 & 89.3 & \textbf{89.9} & \\ 
    \bottomrule
    \end{tabular}
    }
    \caption{AlignScore of language models fine-tuned by different training settings using our approach on the two datasets. The best results are highlighted in \textbf{bold}.}
    \label{tab:main_tab}
\end{table}

\subsection{Similarity of Summary Pairs}
Taking the training outcome of BART on XSUM as the example, we examined the impact of similarity between paired summaries, as shown in Table \ref{tab:random_score}.
Summary pairs generated by selecting alternative outputs, i.e., (BS\#1,BS\#2), achieved higher similarities than pairs generated by varying the decoding strategy, as also shown in Table \ref{tab:main_tab}. 
Highly similar summary pairs help the model focus on subtle factual consistency differences, but we speculate that there could exist a threshold.
The (BS\#1,Greedy) strategy is competitive with (BS\#1,BS\#2) overall in Table \ref{tab:main_tab}, suggesting that an average similarity $\sim0.7$ may be sufficient. 

\begin{table}[htbp]
    \centering
    \tiny
    \begin{tabular*}{0.9\columnwidth}{@{}@{\extracolsep{\fill}}cccc@{}}
    \toprule
    \textbf{Decoding Strategy}& \textbf{Pair Similarity} & \textbf{Method}  & \textbf{AlignScore} \\
    \midrule
    - & - & SFT & 61.9 \\
    \midrule
    \multirow{2}{*}{(BS\#1, BS\#2)} & \multirow{2}{*}{94.0} & MPO & 62.0 \\
    & & Ours & 86.6 \\ 
    \midrule
    \multirow{2}{*}{(BS\#1, Greedy)} & \multirow{2}{*}{82.6} & MPO & 36.3 \\
    & & Ours & 86.1 \\ 
    \midrule
    \multirow{2}{*}{(BS\#1, RS\#1)} & \multirow{2}{*}{34.9} & MPO & 66.4*\\
    & & Ours & 72.0 \\
    \bottomrule
    \end{tabular*}
    \caption{The effect of using different decoding strategies to generate summary pairs for training BART on XSUM. * indicates the result cited from \citet{choi-etal-2024-model}.}
    \label{tab:random_score}
\end{table}

Taking BART as an example, in Table \ref{tab:random_score}, we further investigated the effect of less similar summary pairs (BS\#1,RS\#1), i.e., the best setting for MPO,
to which we applied the same preference label generation process.
Using our method to fine-tune with these labels still improved factuality but to a lesser degree than the similar pairs (BS\#1,BS\#2) and (BS\#1,Greedy). 
Although MPO was able to obtain the largest improvement on (BS\#1,RS\#1) pairs, our method still outperformed it, validating the effectiveness of our method on reducing the noise in the dataset.
Furthermore, we observe the same degeneration mentioned in \citet{choi-etal-2024-model} on the similar pairs (BS\#1,BS\#2) and (BS\#1,Greedy).
We show the evaluation accuracy curve during training in Appendix \ref{app:curve}, which stayed level during training, implying that the model benefitted little from training on these data.
Summary pairs generated by beam search and random sampling, which have a greater factuality gap (as shown in Table \ref{tab:decoding_score}), were too straightforward for BART to learn from, resulting in minimal improvements.

Therefore, we can conclude that both our similar summary pair generation process and our labelling step using automated metrics contribute to the final improvement of our approach.

\begin{figure}[htbp]
    \centering
    \begin{subfigure}{0.4\linewidth}
        \centering
        \includegraphics[width=\linewidth]{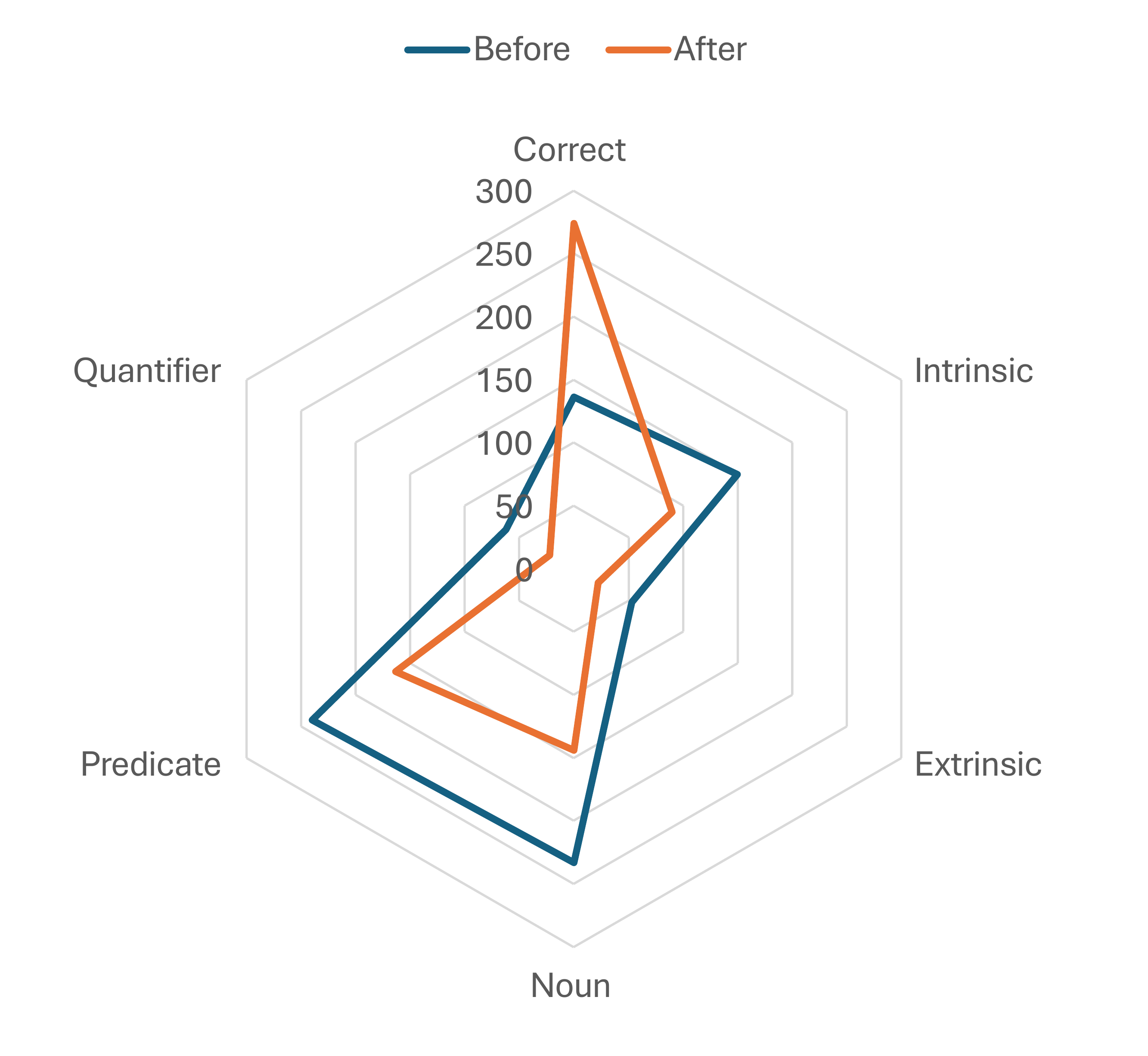}
        \caption{BART}
    \end{subfigure}
    % \vspace{1em} % Optional spacing between the subfigures
    \begin{subfigure}{0.4\linewidth}
        \centering
        \includegraphics[width=\linewidth]{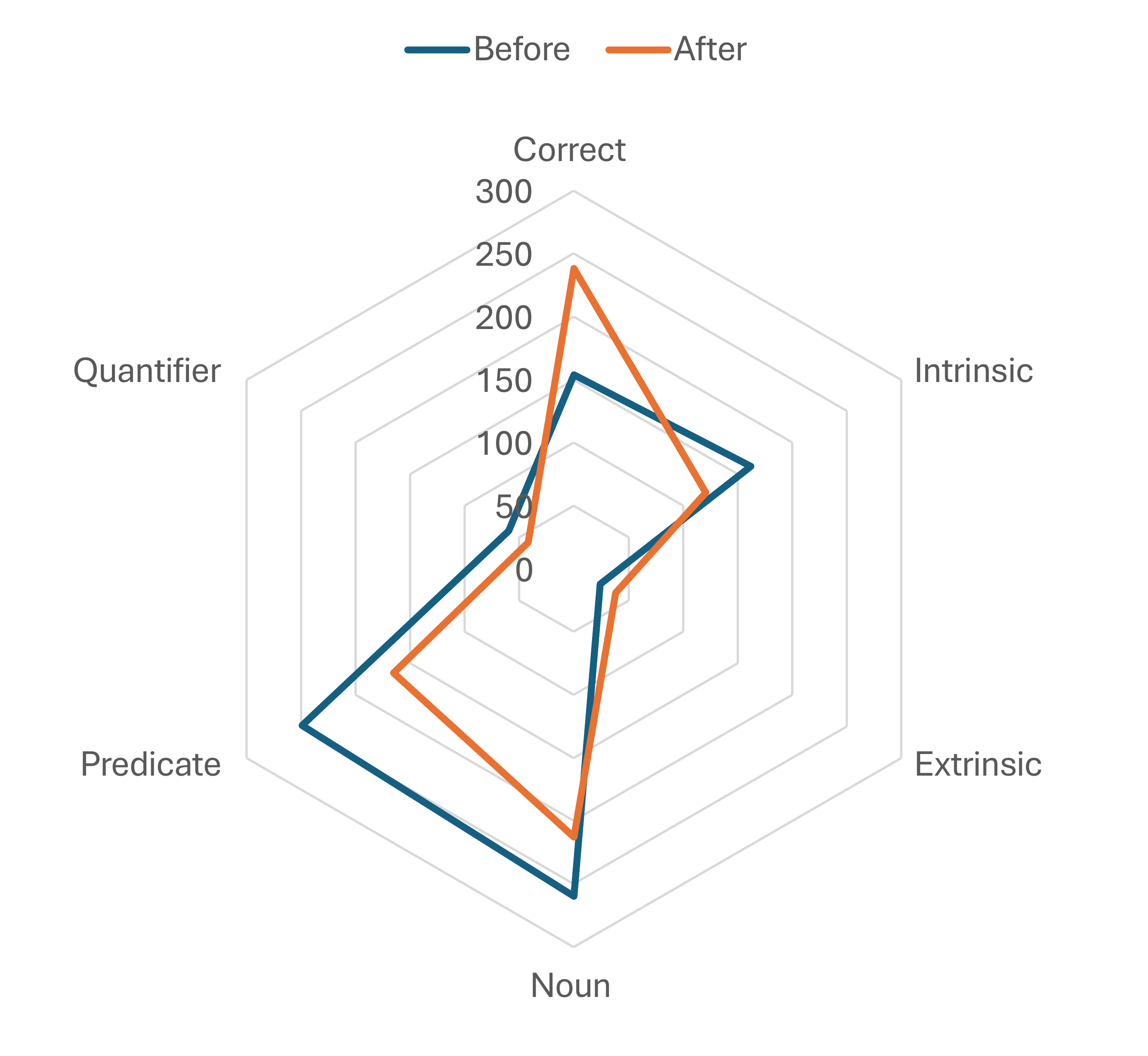}
        \caption{GPT-J}
    \end{subfigure}

        \begin{subfigure}{0.4\linewidth}
        \centering
        \includegraphics[width=\linewidth]{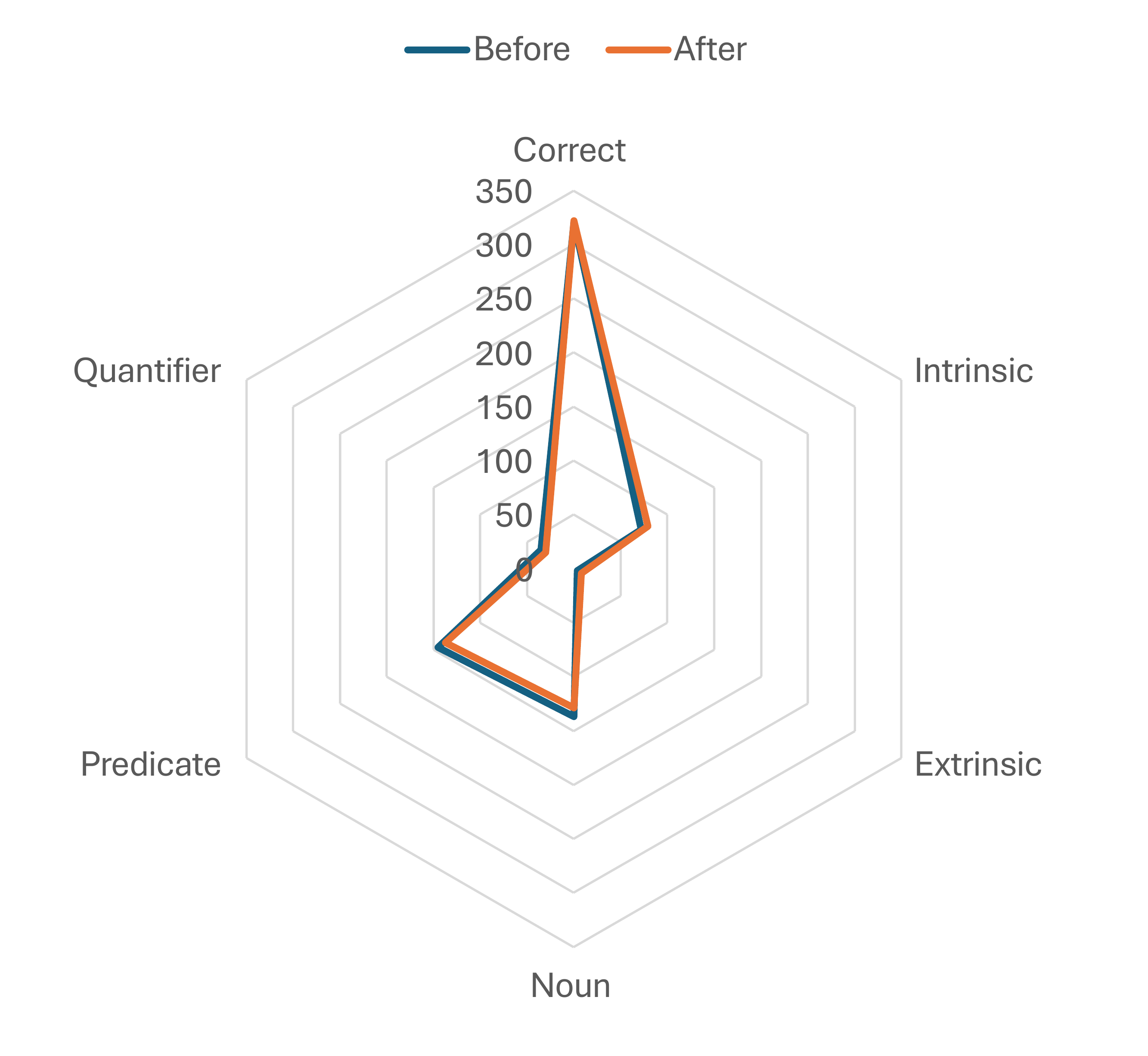}
        \caption{Llama}
    \end{subfigure}
    % \vspace{1em} % Optional spacing between the subfigures
    \begin{subfigure}{0.4\linewidth}
        \centering
        \includegraphics[width=\linewidth]{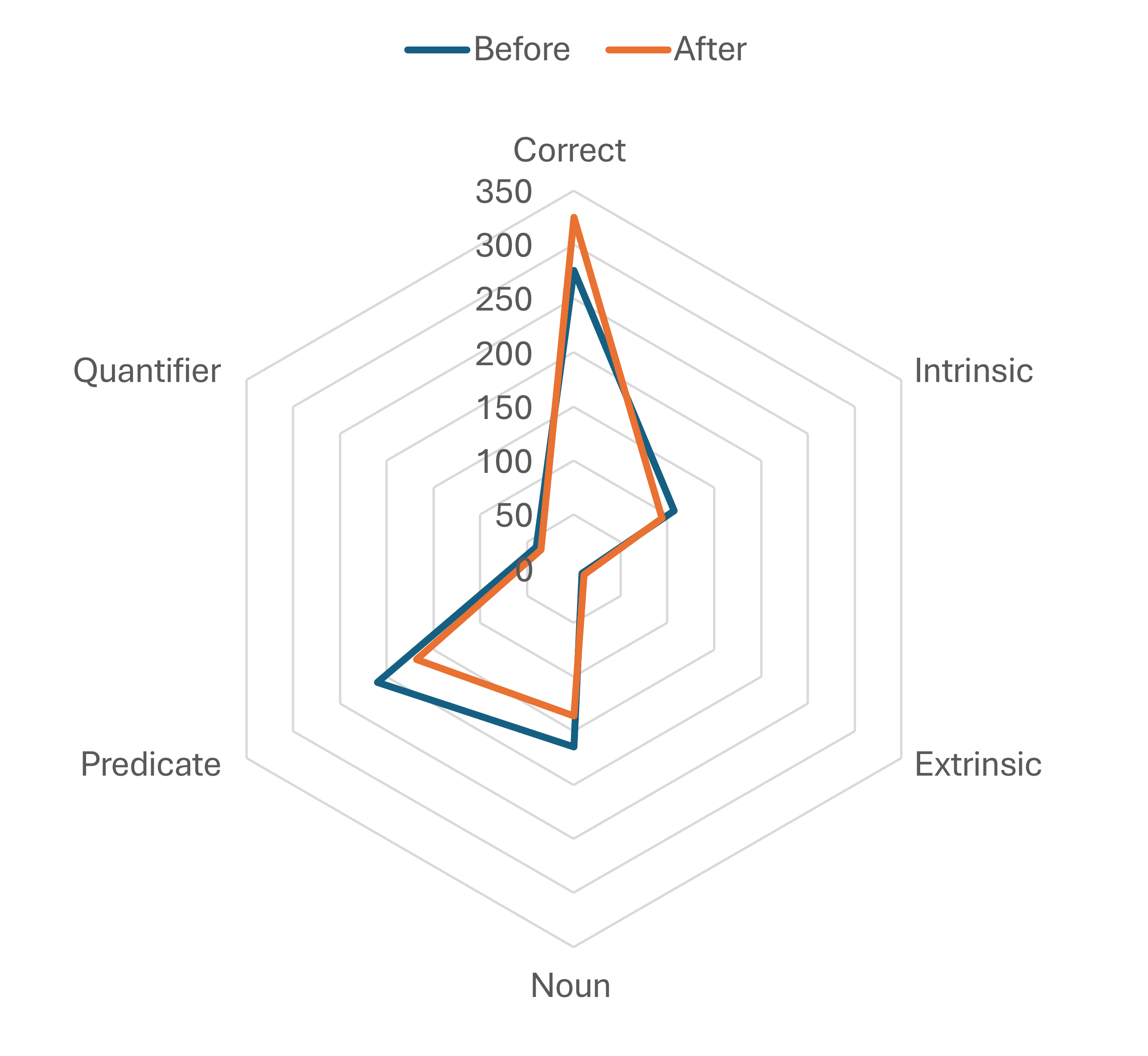}
        \caption{DeepSeek}
    \end{subfigure}
    \caption{Error frequencies on XSUM.}
    \label{fig:error_fq_xsum}
\end{figure}

\begin{figure}[htbp]
    \centering
    \begin{subfigure}{0.4\linewidth}
        \centering
        \includegraphics[width=\linewidth]{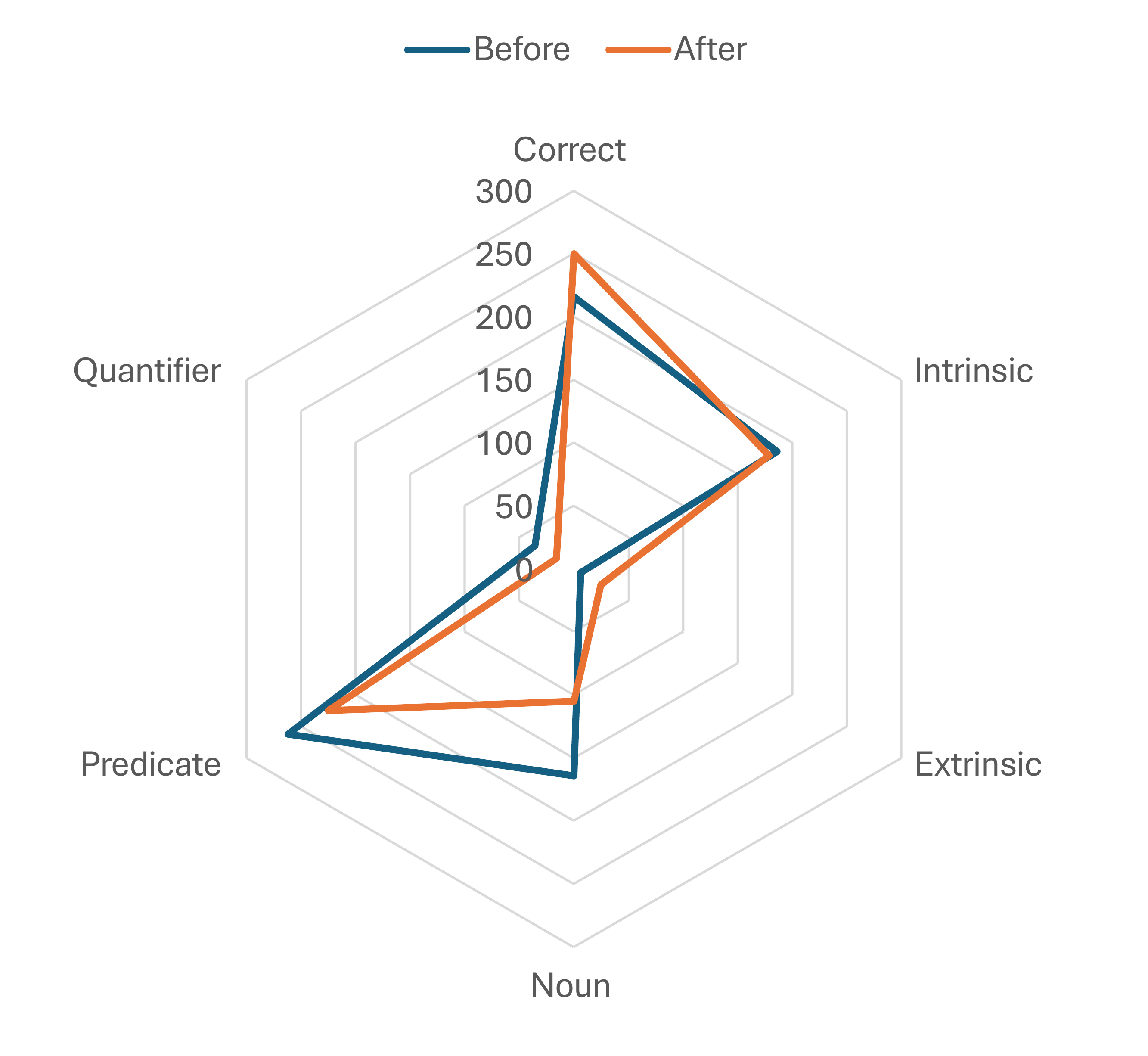}
        \caption{BART}
    \end{subfigure}
    % \vspace{1em} % Optional spacing between the subfigures
    \begin{subfigure}{0.4\linewidth}
        \centering
        \includegraphics[width=\linewidth]{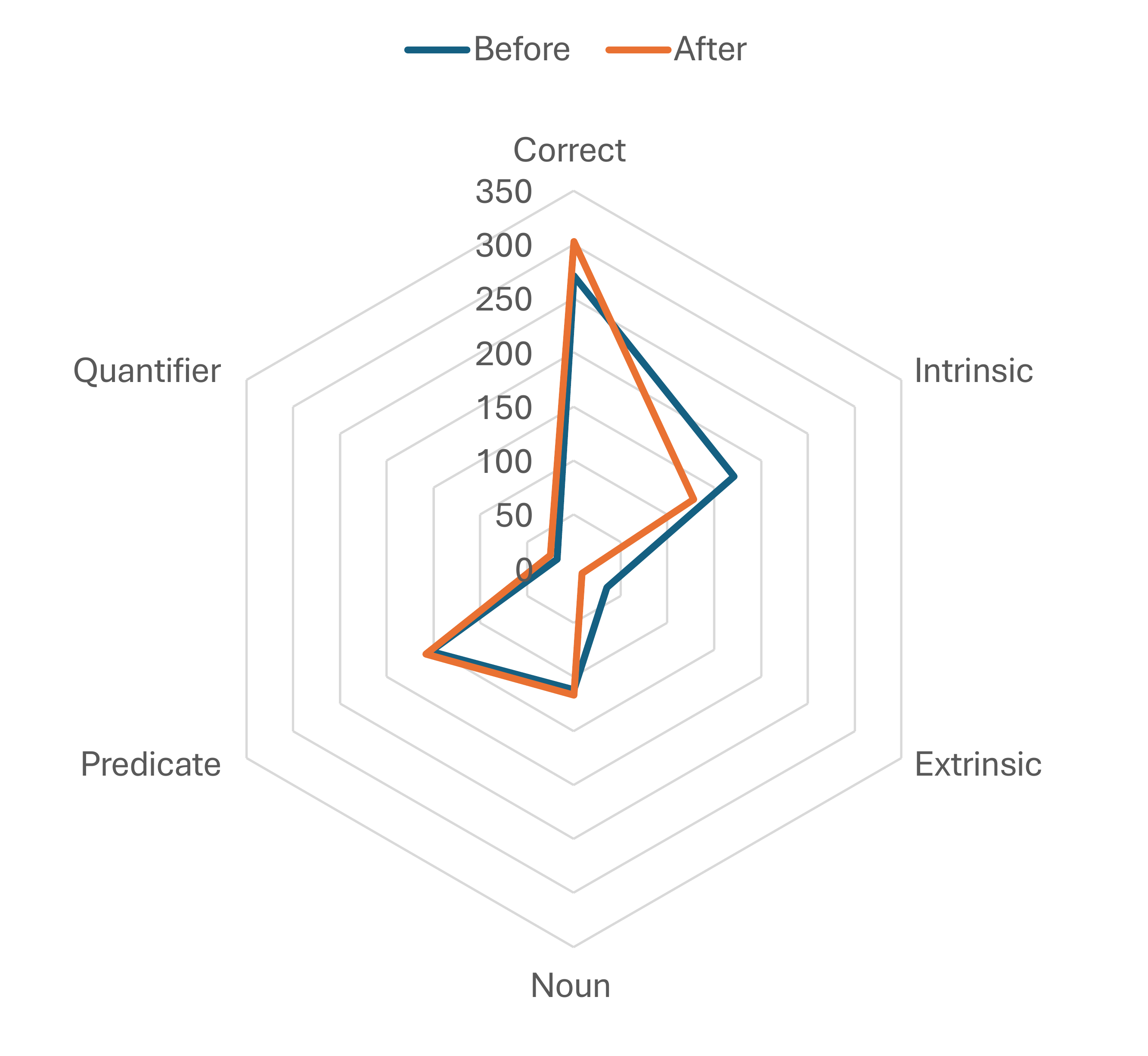}
        \caption{GPT-J}
    \end{subfigure}

        \begin{subfigure}{0.4\linewidth}
        \centering
        \includegraphics[width=\linewidth]{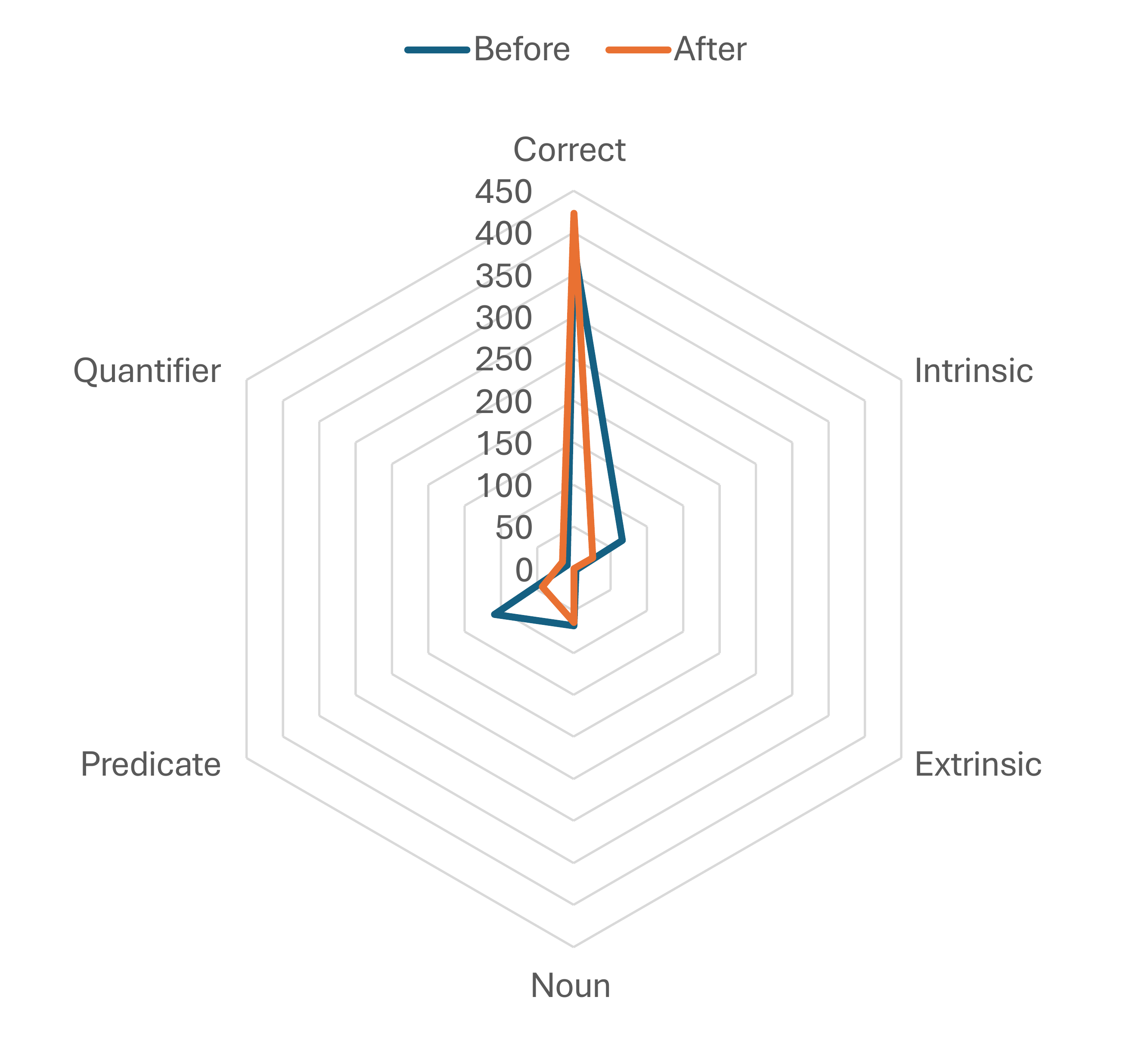}
        \caption{Llama}
    \end{subfigure}
    % \vspace{1em} % Optional spacing between the subfigures
    \begin{subfigure}{0.4\linewidth}
        \centering
        \includegraphics[width=\linewidth]{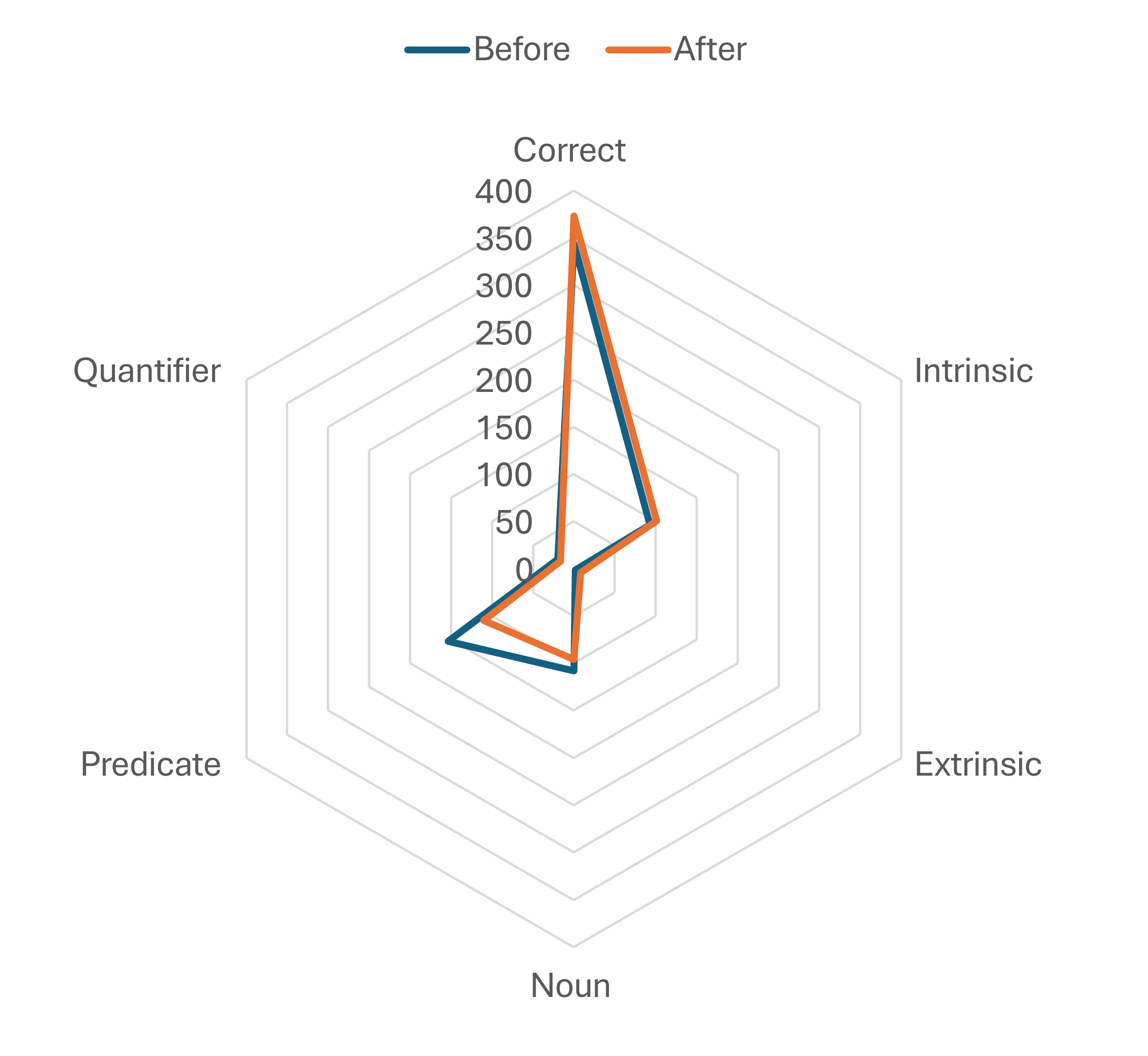}
        \caption{DeepSeek}
    \end{subfigure}
    \caption{Error frequencies on TL;DR.}
    \label{fig:error_fq_tldr}
\end{figure}

\subsection{Disagreement Analysis}
We looked into the rates of each model triggering the disagreement filter on the two datasets. 
In practice, 1000 summaries pairs were generated to obtain the preference labels. 
Table \ref{tab:percentage} below shows that at least 60\% of the data was retained for training after the filtering process.

\begin{table}[htbp]
    \tiny
    \centering
    \begin{tabular}{ccc}
    \toprule
     Model & XSUM & TL;DR \\
     \midrule
     BART & 28.6 & 29.8 \\
     GPT-J & 31.8 & 27.3 \\
     LLaMA & 37.3 & 31.4 \\
     DeepSeek & 34.7 & 28.9 \\
     \bottomrule
    \end{tabular}
    \caption{The percentage of data that triggers disagreement filter in our experiments.}
    \label{tab:percentage}
\end{table}

\subsection{Inconsistency Type Analysis}
Finally, we employ ChatGPT to assess factual inconsistencies in the summaries and analyse how the frequency of factual errors changes before and after training with our approach.
Similar to previous studies \citep{tang-2023-aggrefact}, we defined five inconsistency types, namely \textit{Intrinsic}, \textit{Extrinsic}, \textit{Noun}, \textit{Predicate}, \textit{Quantifier}. 
Along with \textit{Correct} summaries, we asked ChatGPT to identify them according to a given definition and count the frequency of each. 
The definition and prompt can be found in Appendix \ref{app:prompt}.

Figures \ref{fig:error_fq_xsum} and \ref{fig:error_fq_tldr} show that the error frequencies of \textit{Noun}, \textit{Predicate}, and \textit{Quantifier} types mostly decreased. 
Consequently, our approach achieved many more \textit{Correct} summaries than SFT checkpoints, demonstrating the effectiveness of our approach across different models.

\section{Conclusion}
We introduce a novel automatic training pipeline for improving the factual consistency of summarisers.
Our approach can be generalised over different model architectures and scales. 
It requires only source documents, utilising multiple factuality evaluation metrics to score the summary and obtain labels for preference optimisation. 
The experimental results suggest that our approach outperforms baselines and boosts the factuality performance of smaller models to a level comparable to LLMs, revealing the effectiveness of preference learning over similar summary pairs.

\section*{Limitations}
\label{app:limitations}
We only applied SBERTScore and SummaC to score the generated summaries in this paper. 
There are various other metrics available but we were not able to test them all. While we were able to demonstrate that it is possible to improve factuality using our chosen imperfect metrics, this could raise concerns about the generalisation ability of our approach to other automated scoring methods. 

We also did not include results for  RL with the scalar metric scores 
used directly as the reward signal. 
We explored this method in our early investigation and found that it requires both the model and the metric backbone to be large and extremely computationally costly, 
otherwise catastrophic forgetting became a very common problem during training. 
Although we did not have the comparison against this line of work, we believe that our method provides a more stable training paradigm in practice, as we never encountered the catastrophic forgetting problem in our experiments.

In overall quality evaluation, we found that our approach generated summaries that were less preferred by ChatGPT when comparing to SFT/RLHF summaries. 
This reveals the challenge of how to fine-tune the summariser towards better factuality without trading off other qualities. It also highlights the difficulty of judging the overall quality  of summaries, where a human or LLM judge may put more weight on certain qualities (e.g., readability, brevity) at the expense of others (e.g., factual consistency). 
The trade-off between these qualities may need to be judged within the context of a specific application: how important it is that a summary is factually consistent versus stylistically compelling will depend on its use case.

\section*{Acknowledgments}
The authors acknowledge the use of resources provided by the Isambard-AI National AI Research Resource (AIRR). Isambard-AI is operated by the University of Bristol and is funded by the UK Government’s Department for Science, Innovation and Technology (DSIT) via UK Research and Innovation; and the Science and Technology Facilities Council [ST/AIRR/I-A-I/1023]. The financial support for Yuxuan Ye is provided by the programme of the China Scholarship Council (No. 202108060154). Raul Santos-Rodriguez is supported by the UKRI Turing AI Fellowship EP/V024817/1.

\bibliography{latex/anthology,latex/custom}
%\bibliography{custom}
%\newpage
\appendix

%\section{Dataset Characteristics}
%\label{app:dataset}
%In Table \ref{tab:dataset}, 

\section{Evaluation Metrics}
\label{app:metrics}
For a given source document $D$, we denote its generated summary as $S$. 
The length of the source document is denoted $|D|$ and the summary length is denoted $|S|$.

\paragraph{AlignScore} breaks both source document and summary into certain length chunks. We denote the sets of chunks as $\{D_j\}$ and $\{S_i\}$. 
It then aggregates the entailment scores obtained on these text chunk pairs. The backbone NLI classifier is trained for predicting the alignment degree in sentence pairs.
\begin{multline*}
    \text{AlignScore}(S,D)=\\
    \frac{1}{|\{S_i\}|}\sum_{s \in \{S_i\}}\max_{d \in \{D_j\}}{\text{Alignment}(s,d)}
    % {\text{Alignment}(s,d)}
\end{multline*}

\paragraph{FactCC} provides a coarse-grained insight into whether the generated summary is entailed by the source document. The backbone is an NLI model based on BERT, so the equation can be written as
\begin{equation*}
    \text{FactCC}(S,D) = \text{NLI}(S;D)
\end{equation*}

\paragraph{BARTScore} is essentially the weighted log-likelihood computed using BART. It can be written as
\begin{equation*}
    \text{BARTScore}(S,D)=\sum_{t=1}^{|S|}{\omega_{t}\log{p(S_t|S_{<t},D,\Theta)}}
\end{equation*}
where $S_t$ is each token in $S$ and $\Theta$ is the trained weights of BART.

\paragraph{ROUGE} computes the n-gram recall between the candidate and the target text. In this paper, we use ROUGE-L, which match the longest common subsequence (LCS) that appear in both summaries. The equation can be expressed as
\begin{equation*}
    \text{ROUGE-L}(S,D) = \frac{\text{LCS Length}}{\text{Reference Summary Length}}
\end{equation*}

%\section{Model Specifications}
%\label{app:model}

\section{Prompt for LLMs}
\label{app:prompt}
\subsection{Prompt for Summarisation Generation}
\begin{figure}[htbp]
    \centering
    \includegraphics[trim=0 1.3cm 0 0,clip,width=0.75\linewidth]{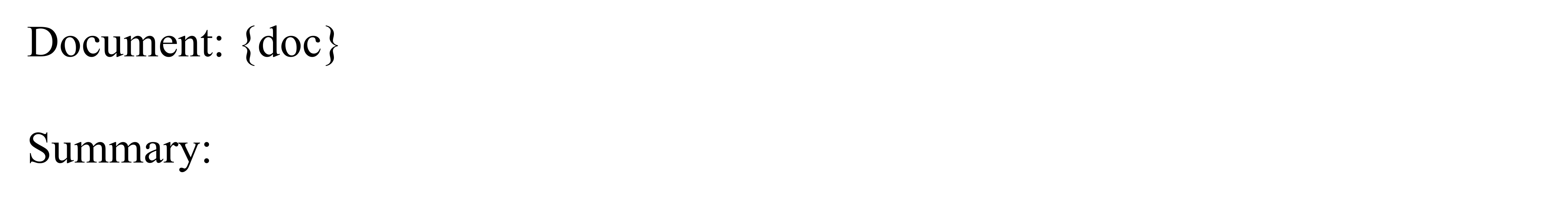}
    \caption{Prompt for GPT-J.}
    \label{fig:gpt-j}
\end{figure}

We only prepare a simple prompt for GPT-J as it needs SFT before applying RL, as shown in Figure \ref{fig:gpt-j}. 
\textit{\{doc\}} denotes the source document which will be changed according to the data being processed. 
It will learn to summarise the source document into a single sentence during SFT, therefore it only needs a template to ensure the model receives the source document and generate summaries as the completion.

\begin{figure}[htbp]
    \centering
    \includegraphics[trim=0 1cm 0 0,clip,width=0.75\linewidth]{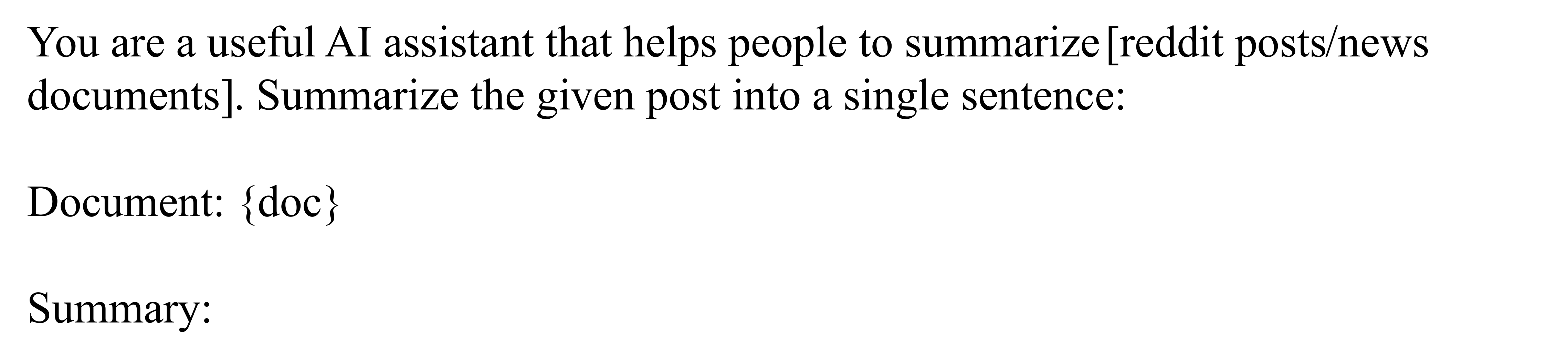}
    \caption{Prompt for LLaMA to generate summaries on the two datasets.}
    \label{fig:llama}
\end{figure}

\begin{figure}[htbp]
    \centering
    \includegraphics[trim=0 1.7cm 0 0,clip,width=0.75\linewidth]{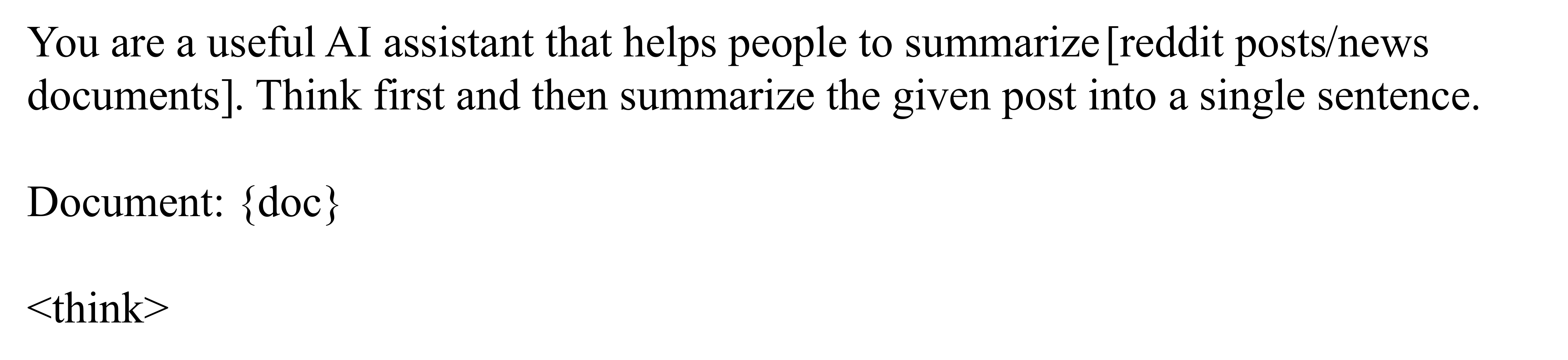}
    \caption{Prompt for DeepSeek to generate summaries on the two datasets.}
    \label{fig:deepseek}
\end{figure}

Figure \ref{fig:llama} and \ref{fig:deepseek} present the prompts we used to generate summaries using LLaMA and DeepSeek on the two datasets. 
The only difference in the prompt is that we indicate that the source documents are reddit posts in TL;DR and news documents in XSUM.

DeepSeek requires a special token \textit{<think>} to trigger the thinking process, as shown in Figure \ref{fig:deepseek}.
Following the prompt, it generates a chain-of-thought that ends with \textit{</think>} before generating the final output.
Therefore, we take all the output after \textit{</think>} as the final summary for the metrics to score.

\subsection{Prompt for ChatGPT Evaluation}
\begin{figure}[htbp]
    \centering
    \includegraphics[trim=0 1.7cm 0 0,clip,width=0.75\linewidth]{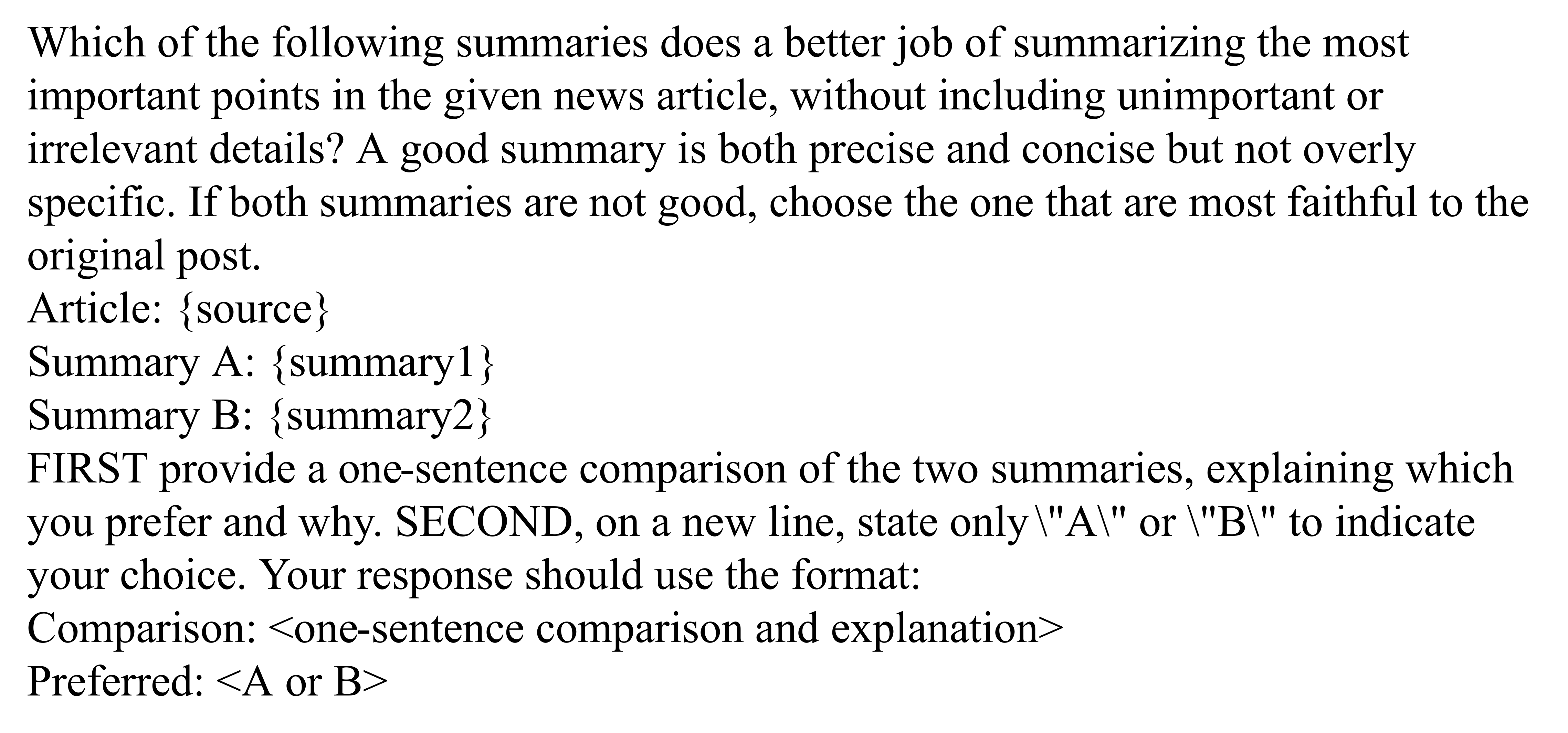}
    \caption{Prompt for ChatGPT win rate evaluation.}
    \label{fig:chat-win-rate}
\end{figure}

We use a similar prompt to the previous work \citep{choi-etal-2024-model} for ChatGPT to compare two summaries, as described in Figure \ref{fig:chat-win-rate}. 
\textit{\{source\}}, \textit{\{summary1\}}, \textit{\{summary2\}} denote the source document and two candidate summaries. 
We found that ChatGPT-4o-mini tends to claim that both summaries are not good enough due to informativeness, therefore we relaxed the requirement and ask it to choose the most faithful summary if both are not good as we focus on factuality on this paper.

\begin{figure}[htbp]
    \centering
    \includegraphics[trim=0 1.7cm 0 0,clip,width=0.75\linewidth]{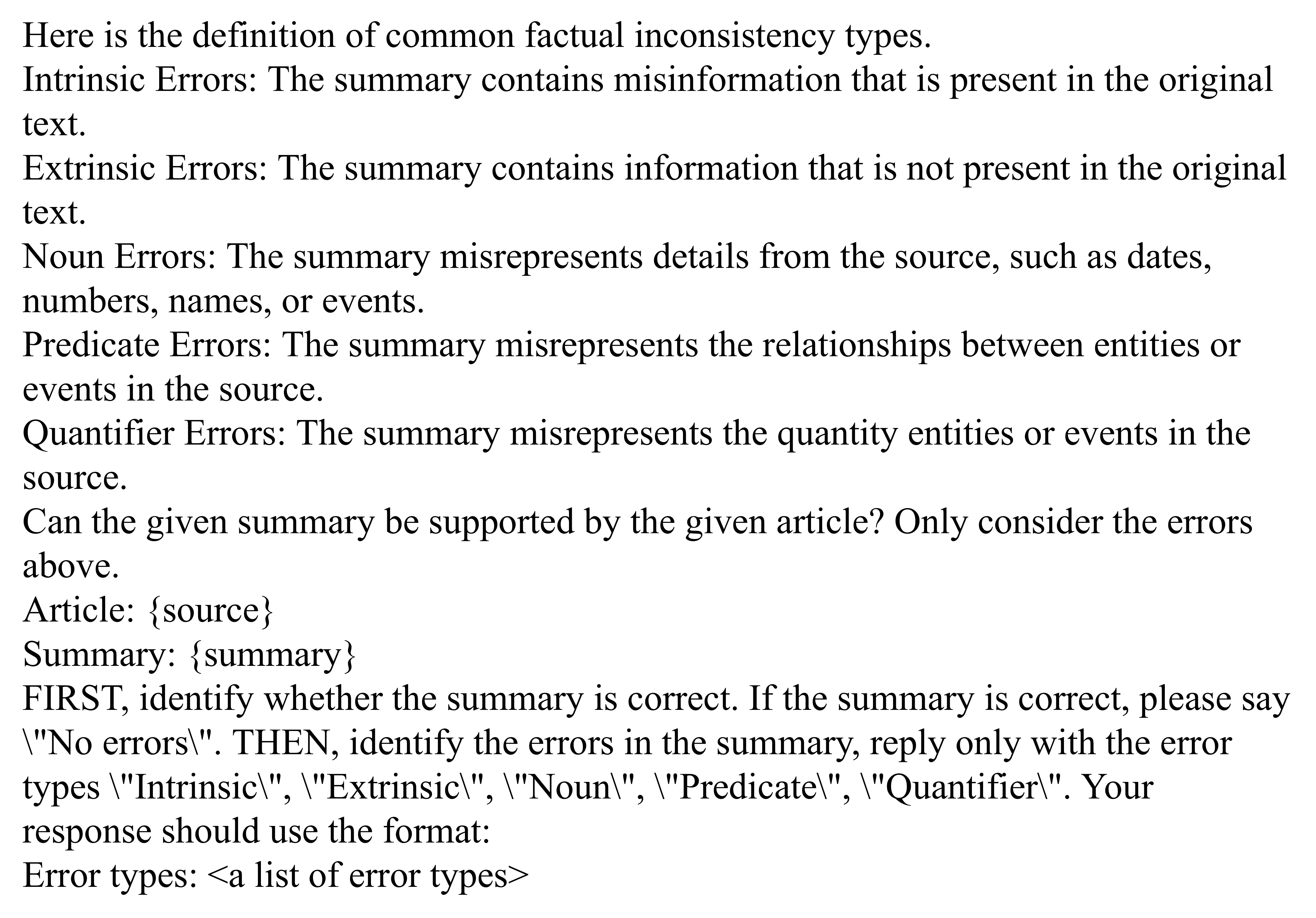}
    \caption{Prompt for ChatGPT inconsistency type analysis.}
    \label{fig:chat-error}
\end{figure}

As for inconsistency type analysis, we give the definition in the prompt first and then ask ChatGPT to judge the summary. 
The prompt is shown in Figure \ref{fig:chat-error}. 
\textit{\{source\}} and \textit{\{summary\}} represent the source document and the summary to analyse. 

\section{Scoring Metric Definitions}
\label{app:def}
For a given source document $D$, we denote its generated summary as $S$. Their sentence collections are marked as $\{D_j\}$ and $\{S_i\}$ respectively. 
SBERTScore is defined as below.
\begin{multline*}
    \text{SBERTScore}(S,D)=\\
    \frac{1}{|\{S_i\}|}\sum_{s \in \{S_i\}}\max_{d \in \{D_j\}}{\text{cos sim}(s,d)}
\end{multline*}

SummaC firstly computes NLI scores on sentence pairs to get a score matrix $A$, such that
\begin{equation*}
\begin{split}
    A_{ij}=\text{NLI}(s,d) \quad & s \in \{S_i\},d \in \{D_j\}
\end{split}
\end{equation*}
It then maps $A$ into a score frequency matrix $H=\text{bin}(A)$, where it bins the NLI scores into $h$ evenly spaced bins for each summary sentence. 
Then a convolutional layer is trained to aggregate $H$ to obtain the final score.
\begin{equation*}
    \text{SummaC}(S,D)=\text{Conv}(H)
\end{equation*}

\section{ChatGPT Win Rate Reason Analysis}
\label{app:reason}
\begin{figure}[htbp]
    \centering
    \includegraphics[width=0.5\linewidth]{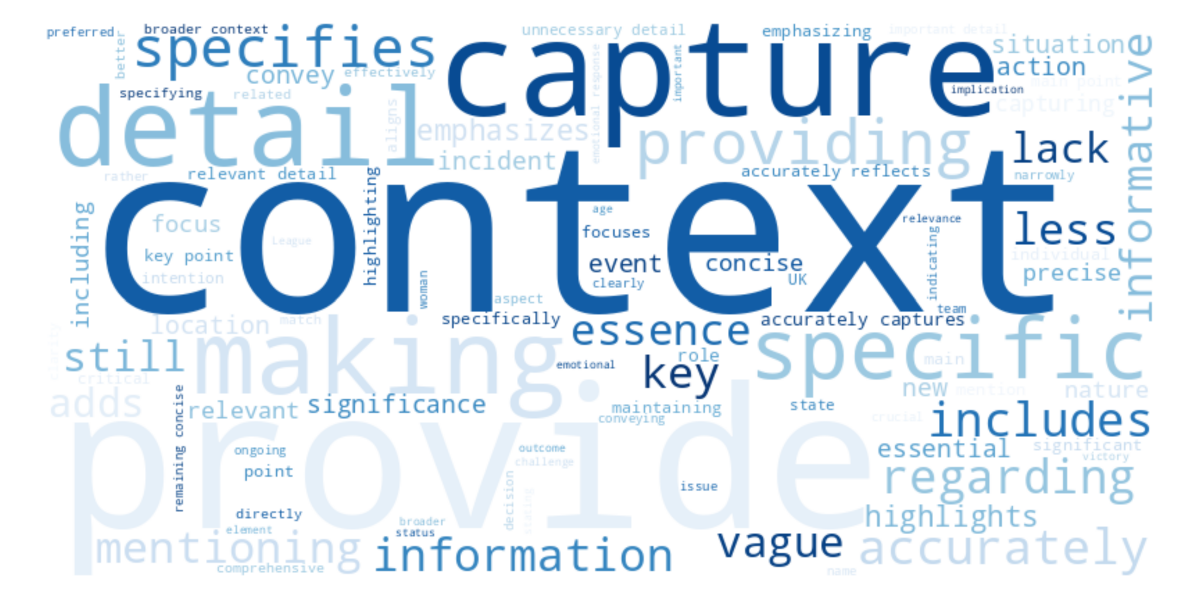}
    \caption{Word cloud showing frequency of terms in the reasons generated by ChatGPT for preferring SFT and RLHF summaries over those produced by our approach.}
    \label{fig:word_cloud}
\end{figure}

We print out the common words that appeared in the reasons given by ChatGPT for choosing SFT and RLHF summaries over ours in Figure \ref{fig:word_cloud}. 
The main reason for the SFT and RLHF summaries being preferred is that they carry more details, while ours reduced the hallucination risk by generating fewer details.

\section{Evaluation Accuracy Curve during Training}
\begin{figure}[htbp]
    \centering
    \includegraphics[width=0.7\columnwidth]{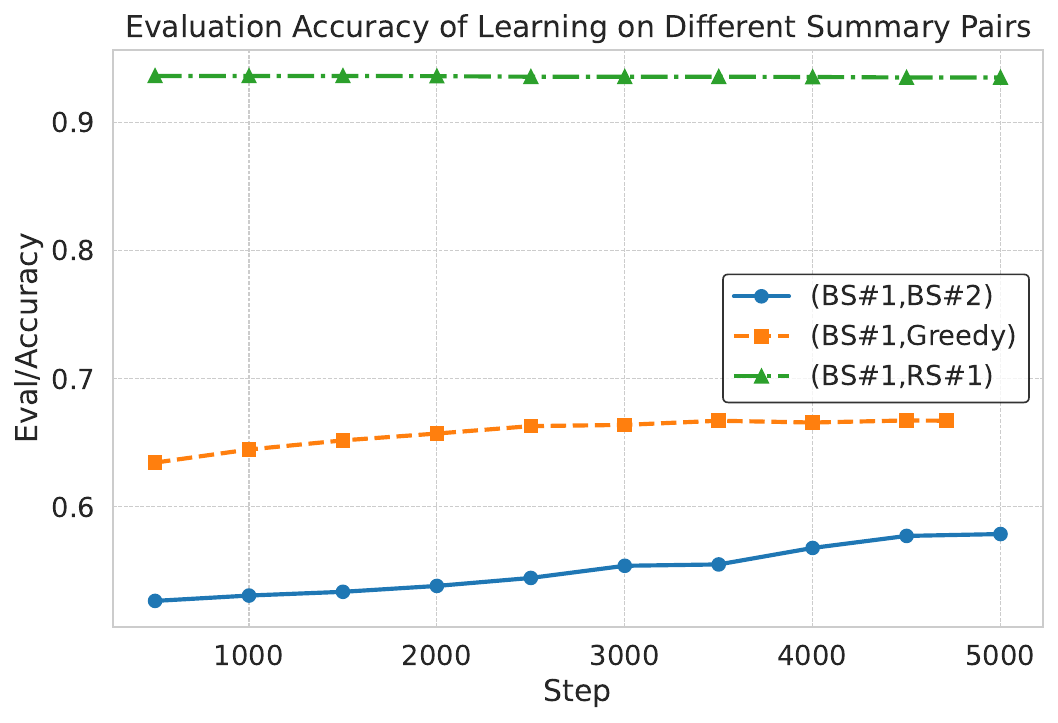}
    \caption{Evaluation accuracies over pairwise labels during DPO training for BART on XSUM. }
    \label{fig:training_curve}
\end{figure}
\label{app:curve}
Figure \ref{fig:training_curve} shows how well the model learns to distinguish the chosen summary and the rejected summary in the pair. 
Ideally, the model learns to simulate the chosen summary while differs its behaviour from the rejected summary so that it gains better accuracies during training.

\end{document}